# Dual-View Selective Instance Segmentation Network for Unstained Live Adherent Cells in Differential Interference Contrast Images

Fei Pan[†1], Yutong Wu[†2], Kangning Cui[1,2], Shuxun Chen[3], Yanfang Li[3], Yaofang Liu[1,2], Adnan Shakoor[4], Han Zhao[3], Beijia Lu[2], Shaohua Zhi[5], Raymond Chan[1,2], and Dong Sun[*3]

[1]Hong Kong Centre for Cerebro-Cardiovascular Health Engineering, Hong Kong, China
[2]Department of Mathematics, City University of Hong Kong, Hong Kong, China
[3]Department of Biomedical Engineering, City University of Hong Kong, Hong Kong, China
[4]Control and Instrumentation Department, King Fahd University of Petroleum and Minerals, Dhahran, Saudi Arabia
[5]Department of Health Technology and Informatics, the Hong Kong Polytechnic University, Hong Kong, China

January 26, 2023

## Abstract

Despite recent advances in data-independent and deep-learning algorithms, unstained live adherent cell instance segmentation remains a long-standing challenge in cell image processing. Adherent cells' inherent visual characteristics, such as low-contrast structures, fading edges, and irregular morphology, have made it difficult to distinguish from one another, even by human experts, let alone computational methods. In this study, we developed a novel deep-learning algorithm called dual-view selective instance segmentation network (DVSISN) for segmenting unstained adherent cells in differential interference contrast (DIC) images. First, we used a dual-view segmentation (DVS) method with pairs of original and rotated images to predict the bounding box and its corresponding mask for each cell instance. Second, we used a mask selection (MS) method to filter the cell instances predicted by the DVS to keep masks closest to the ground truth only. The developed algorithm was trained and validated on our dataset containing 520 images and 12198 cells. Experimental results demonstrate that our algorithm achieves an $AP^{segm}$ of 0.555, which remarkably overtakes a benchmark by a margin of 23.6%. This study's success opens up a new possibility of using rotated images as input for better prediction in cell images.

## 1 Introduction

The cell, the fundamental unit of life, is a complex of material metabolism, energy conversion, and information regulation. For a typical cell, whether a bacterial or an animal cell, water accounts for about 70% of its weight, which causes it transparent [1]. Consequently, when such a cell is observed under a bright-field microscope, the contrast is very weak, leading to poor image quality. So, it is best to use a phase contrast microscope or a differential interference contrast (DIC) microscope to observe live cells. The former, a phase contrast microscope, reveals more detail of a cell's internal structures and discerns its attachments to nearby cells. While the latter, a DIC microscope, provides pseudo-three-dimensional images with a shadow-cast appearance.

In addition to these two imaging modes, fluorescence microscopy is a commonly used approach for observing specific macromolecules, such as proteins and nucleic acids in cells in modern biological laboratories [2]. In a fluorescence microscope, a short-wavelength excitation light passing through the excitation filter irradiates the fluorescent molecules (fluorophores) marked in the sample to generate visible light of a particular wavelength that can be seen by the viewer or digitally captured using a complementary metal oxide semiconductor (CMOS) or charge-coupled device (CCD). However, fluorescence microscopy also brings several disadvantages, such as photo-bleaching and photo-toxicity, so unstained microscopy is still the most common non-invasive approach for observing live cells [3].

---

*Corresponding Author: medsun@cityu.edu.hk; [†] The two authors contributed equally to this work.

Concurrent with progress in optics and advances in imaging, cell image processing [4] has been in increasing demand in biomedical research. Typical tasks in cell image processing include image classification, image segmentation, object tracking, and augmented microscopy [5]. Here cell detection is a primary task, aiming to locate each cell's positions using a bounding box. In contrast, cell instance segmentation is a more demanding task that aims at detecting each instance of different cells and generates its segmentation mask, even if they are of the same class in an image. Inaccuracies of cell instance segmentation can bring extensive consequences for diverse downstream applications, such as cell culture characteristics estimation [6], cell micromanipulation [7, 8], digital pathology [9, 10], and computer-aided diagnosis (CAD) [11, 12].

Although several convolutional neural networks (CNNs) [13–17] and relevant cell image datasets [18, 19] have been proposed recently to solve this problem under various imaging circumstances, accurate instance segmentation of unstained live adherent cells in DIC images—a common situation in many biomedical experiments—remains unsolved. For computational researchers, this is mainly due to the lack of established datasets; but for biomedical researchers, it is primarily due to the lack of accurate and out-of-the-box algorithms. More specifically, the difficulty of instance segmentation for unstained adherent cells lies in four aspects, illustrated in Figs. 1 and 2. First, adherent cells' morphology and orientations are heterogeneous. Second, sporadic individual cells' edges usually fade into the image background. Third, a few cells are sick or dying, exhibiting unusual features. Fourth, adherent cells often gather together and thus make their bordering edges indistinguishable. These characteristics pose a prohibitive barrier in manual annotations, let alone establishing a high-quality dataset. Both early data-independent and recent deep-learning algorithms are primarily centered around fixed and stained histopathological images. As such, this study aims to fill the gap by providing a new instance segmentation algorithm dual-view selective instance segmentation network (DVSISN) for unstained live adherent cells in DIC images.

Without bells and whistles, DVSISN surpasses several major state-of-the-art (SOTA) CNNs on our dataset. Particularly, DVSISN achieves 0.634 in $AP^{bbox}$ and 0.555 in $AP^{segm}$, approximately 10% to 20% better than its counterparts. Such an improvement is made by the following two methodological innovations in this study. (1) A dual-view segmentation (DVS) method is proposed to take combinations of original and rotated input images to increase the coverage of bounding boxes. (2) An mask selection (MS) method is proposed to keep the finest masks in a supervised way.

The rest of this article is organized as follows. Section 2 briefly introduces image segmentation and the region-based convolutional neural network (R-CNN) family. Section 3 describes our cell image dataset. Section 4 gives details of DVSISN. Section 5 reports quantitative comparisons of DVSISN against other SOTA algorithms. Finally, Section 6 concludes the study and discusses future work.

## 2 Related Work

### 2.1 Image Segmentation

Image segmentation is partitioning an image into multiple segments or components. Depending on the complexity of the task, image segmentation can be divided into three main categories: 1) semantic segmentation, that is, classifying pixels with semantic labels; 2) instance segmentation, that is, identifying and segmenting individual objects; and 3) panoptic segmentation, that is, unifies semantic & instance segmentation [20].

Semantic segmentation, also called scene labeling, predicts semantic labels for each pixel in an image. It has been a critical task in computer vision for decades, for which researchers have developed methods ranging from thresholding to SOTA CNNs [21–23]. These techniques are widely used in many applications, such as autonomous driving [24], remote sensing [25, 26], and medical image processing [27, 28].

In recent years, instance segmentation has become one of computer vision's most critical and problematic directions. Other than semantic segmentation, instance segmentation identifies and segments all instances in an image belonging to different categories. Existing R-CNN-based instance segmentation algorithms need two stages in general, i.e., detecting bounding boxes that contain objects and then predicting foreground masks for each region of interest (RoI) [29, 30]. In contrast, other one-stage approaches adopt fully convolutional models for instance segmentation without an explicit feature localization, such as YOLACT [31]. Instance segmentation benefits applications in many fields, like robotics [32] and autonomous driving [33].

Cell instance segmentation is strongly needed in biomedical applications, for example, cell micromanipulation [7, 8], digital pathology [9, 10], and CAD [11, 12]. Cell instance segmentation aims to separate each cell instance and predict its corresponding class in input images. Although it has been a complicated task for a long time, several deep learning algorithms were proposed recently to increase the accuracy and robustness [13–16].

## 2.2 R-CNN Family

R-CNN [34] is the first algorithm to successfully apply deep learning to object detection. It can be divided into three main steps: (1) extracting and wrapping region proposals from each image, (2) computing CNN features for each warped patch, and (3) classifying each region and deleting redundant predictions.

Despite its breakthrough advances, R-CNN still has several drawbacks, for example, high computational cost and multi-stage tuning. To further improve the efficacy of R-CNN, Fast R-CNN [35] feeds the whole image into a CNN to extract features and uses selective search [36] to reduce repeat computations. The wrap step is replaced by spatial pyramid pooling [37] to avoid distortions. Besides, Fast R-CNN adopts a multi-task loss to train the softmax classifier and the bounding box regressor end-to-end. Later, Faster R-CNN [38] applies a region proposal network (RPN) to optimize the quality of region proposals with lower computational costs. Rotated Cascade R-CNN [39] incorporates rotated bounding boxes to detect quadrangular and curved objects efficiently.

Later, Mask R-CNN [29] extended Faster R-CNN by adding a mask branch to achieve instance segmentation. Mask R-CNN uses the RPN for each input image to search RoIs as Faster R-CNN did. Then the class and box offset are predicted for each RoI; in parallel, a binary mask that encodes the spatial layout of the contained object is generated. Soon, Mask Scoring R-CNN (MS R-CNN) [40] improves the inconsistency between the classification and binary mask quality of Mask R-CNN by adding a network block to compute the mask score. Cascade R-CNN [41] uses a multi-stage architecture based on Faster R-CNN that is trained with increasing intersection over union (IoU) thresholds stage by stage to balance the trade-off between performance and IoU threshold setting. Recently, Rotated Mask R-CNN has adopted a rotated bounding box representation to enhance the performance of Mask R-CNN on dense objects [42].

## 3 Dataset

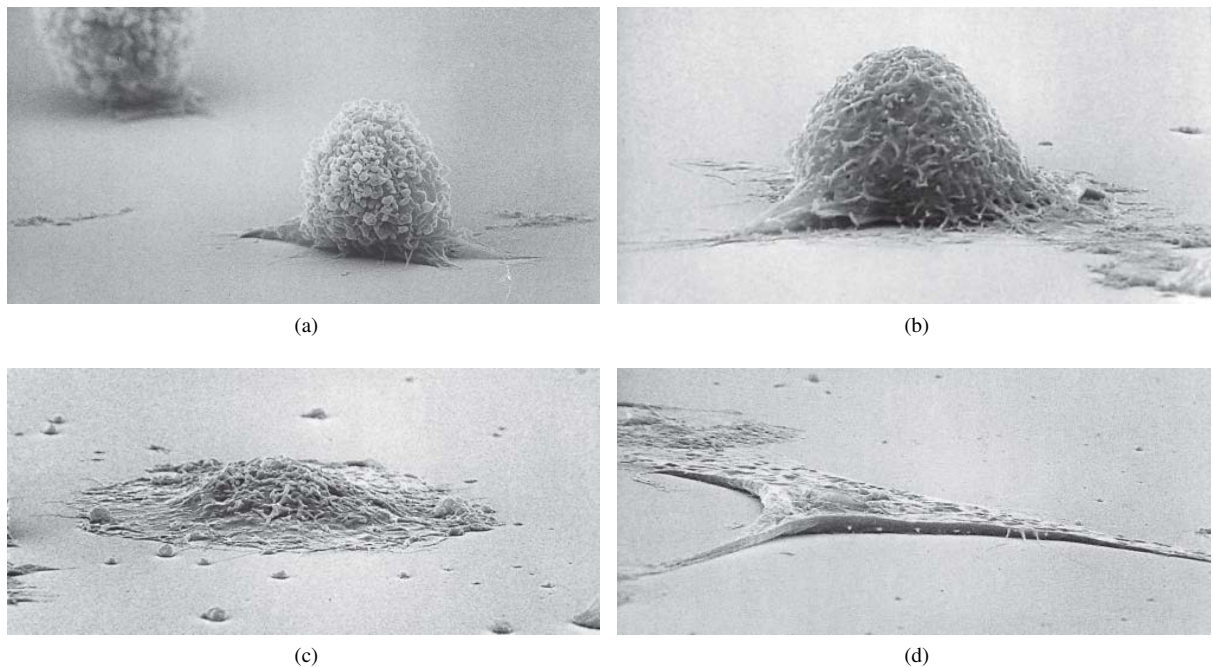

Fig. 1: The morphology of Swiss 3T3 mouse fibroblasts in 4 steps in the cell-spreading process on glass coverslips. Cells were fixed and shown after (a) 30 minutes, (b) 60 minutes, (c) 2 hours, and (d) 24 hours of attachment. SOURCE: Jonathan J. Rosen and Lloyd A. Culp, *Exp. Cell Res.* 107:141, 1977 [43] with permission from Elsevier.

Most cells derived from vertebrates, such as birds and mammals, except for hematopoietic cells, germ cells, and a few others, are adherent cells. Adherent cells, as opposed to suspension cells, are anchorage-dependent and must be cultured on a tissue-culture-treated substrate to allow cell adhesion and spreading, as shown in Fig. 1. From the perspective of morphology, adherent cells can be classified into fibroblast-like and epithelial-like cells. The former is bipolar or multi-polar and usually has elongated shapes, while the latter is polygonal and grows as discrete patches [44]. Both cells have a highly irregular morphology compared with the spherical shape of suspensions cells, bringing considerable difficulties for an algorithm to detect, segment, track, and analyze [3, 5, 45].

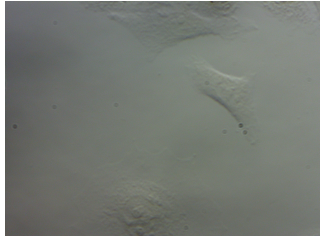

(a-1) A DIC image of sparsely distributed adherent cells.

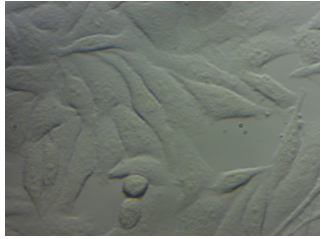

(b-1) A DIC image of densely distributed adherent cells.

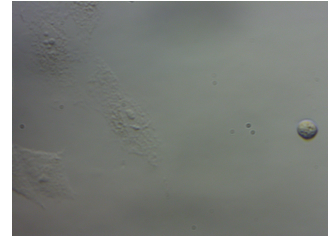

(c-1) A DIC image of unhealthy adherent cells.

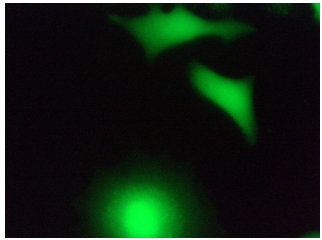

(a-2) A fluorescence image of stained cells in Fig. 2(a-1).

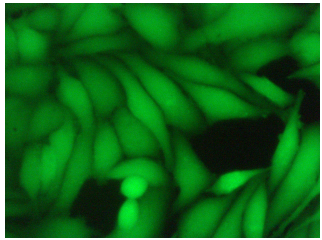

(b-2) A fluorescence image of stained cells in Fig. 2(b-1).

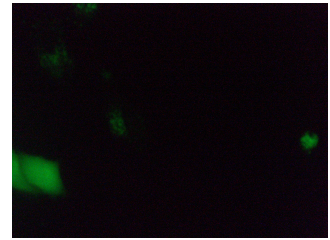

(c-2) A fluorescence image of stained cells in Fig. 2(c-1).

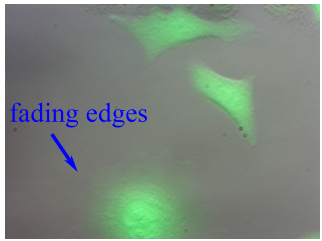

(a-3) A merged image of Figs. 2(a-1) and 2(a-2).

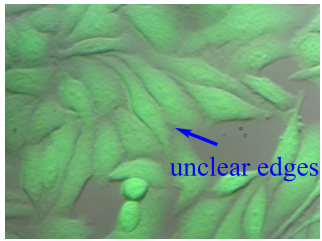

(b-3) A merged image of Figs. 2(b-1) and 2(b-2).

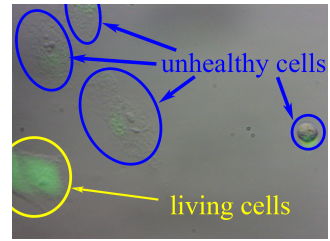

(c-3) A merged image of Figs. 2(c-1) and 2(c-2).

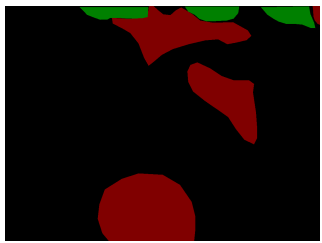

(a-4) Annotated cell classes of Fig. 2(a-1).

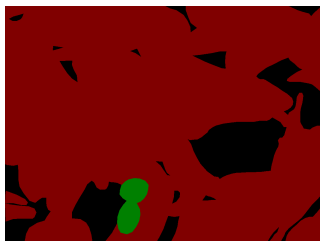

(b-4) Annotated cell classes of Fig. 2(b-1).

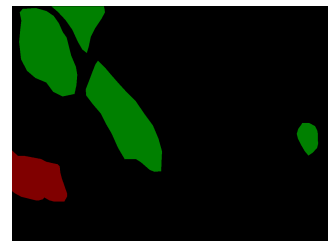

(c-4) Annotated cell classes of Fig. 2(c-1).

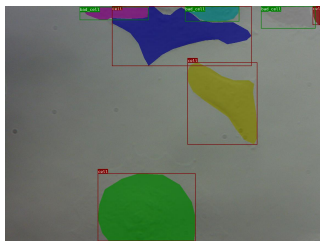

(a-5) Annotated ground truth of Fig. 2(a-1).

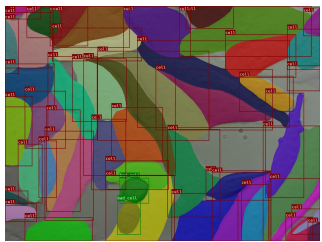

(b-5) Annotated ground truth of Fig. 2(b-1).

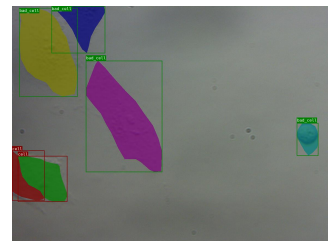

(c-5) Annotated ground truth of Fig. 2(c-1).

Fig. 2: Representative microscopic images of HepG2 human liver cancer cells and their annotated ground truth.

Nevertheless, adherent cells are transparent, so can hardly be observed under a light microscope unless stained. Conventionally, researchers usually use DIC microscopes because they can observe delicate structures in live or unstained specimens and render three-dimensional images with a sense of relief. The working principle is that a DIC microscope converts the phase difference of the object into amplitude changes through the interference of coherent light beams between which the distance is relatively small, only 1 µm or less, inside and outside the sample.

This study used HepG2 human liver cancer cells to provide cell images. Cells were cultured in the Dulbecco's modified eagle medium (DMEM) (Gibco) supplemented with 10% fetal bovine serum (FBS) (Gibco), 100 U/mL of penicillin, and 100 U/mL of streptomycin in a 35 mm glass-bottom Petri dish (culture dish 801002, Wuxi NEST Biotechnology) and placed in a humidified atmosphere of 37 °C and 5% $CO_2$. Calcein acetoxymethyl (AM), a commonly used fluorescent dye, was used to test cell viability and for short-term staining. Before image collection, 5 µL 4 mmol calcein AM (L6037S, US Everbright Inc.) was taken from the refrigerator and restored to room temperature. Then it was mixed with 10 mL phosphate-buffered saline (PBS) to stain the cultured cells. Because the calcein AM emits 530 nm fluorescence when excited by a 488 nm laser, live cells stained with the calcein AM look green.

After cell staining, the dish was transferred from the incubator to the inverted fluorescence microscope (Eclipse Ts2R-FL, Nikon). The microscope was equipped with a motorized XY stage (ProScan H117P1N4, Prior Scientific) and a CMOS camera (DigiRetina 16, Tucsen Photonics). A homemade control software [7, 46] first drove the motorized stage to move the dish (and the cultured cells) to predefined locations to capture DIC images and then drove the stage again to move the dish to the same locations to capture fluorescence images. 520 pairs of DIC and fluorescence images were captured under a $40\times$ objective lens (CFI S Plan Fluor ELWD 40XC 228 MRH08430, Nikon). All images were RGB color images and resized to 1152 pixel × 863 pixel, representing approximately 216.500 µm × 162.375 µm in the dish. Each pair of a DIC image [Figs. 2(a-1) to 2(c-1)] and its fluorescence counterpart [Figs. 2(a-2) to 2(c-2)] is merged for manual annotation [Figs. 2(a-3) to 2(c-3)]. Annotated images [Figs. 2(a-5) to 2(c-5)] can be read by the labelme software [47].

Adherent cells can be roughly classified into two types from the perspective of cell health: healthy (live) and unhealthy (dead or loosely attached) cells, as indicated in Fig. 2(c-3). Healthy adherent cells usually adhere to the culture surface, having irregular morphology and looking completely green in the fluorescence images once stained by the calcein AM. As a comparison, some unhealthy cells, for example, dead cells, can hardly be stained by the calcein AM and only look negligibly green. Other unhealthy cells, though, can be successfully stained by the calcein AM but loosely adhere to the culture surface and are not ideal candidates for typical biomedical experiments, such as cell microinjection.

In addition to classifying cells by how healthy they are, they can also be classified by how densely they grow, as shown in Figs. 2(a-1) and 2(b-1). Sparsely distributed cells [Fig. 2(a-1)] are relatively easy to recognize, but densely distributed cells [Fig. 2(b-1)] are difficult to be distinguished from one another even by humans, so only by live cell staining [Fig. 2(b-2)], can people distinguish individual cells clearly [Figs. 2(b-3) and 2(b-5)].

## 4 Dual-View Selective Instance Segmentation Network (DVSISN)

Since adherent cells are elongated, often tightly closed to one another, and frequently at oblique angles. A natural doubt is that merely a horizontal bounding box cannot capture a sloping cell without including its adjacent cells, thus making predicting masks harder. For example, Mask R-CNN predicts binary masks for each RoI using a fully convolutional network (FCN) [22] that is sufficient for segmenting scattered objects. However, a preliminary experiment reveals that Mask R-CNN produces duplicate predictions of RoIs and causes the predicted masks of cell edges to be vague. Consequently, an intuitive question is that can we apply a rotation operation of 45° on input images before data augmentation?

Fig. 3 shows an overview of our instance segmentation algorithm for adherent cells, a two-part trainable CNN. Its first part is a DVS responsible for producing binary segmentation masks with class labels. Its second part is an MS in charge of removing redundant cell instances and keeping the finest ones. Details of our algorithm are elucidated as follows.

### 4.1 Dual-View Segmentation (DVS)

The DVS part extends the structure of Mask R-CNN, as shown in Fig. 3 (left). First, we augment each input image by rotating it by 45° and then pass the two views of the image to the backbone for feature extraction. RPN is then applied to generate region proposals from the extracted feature maps. RoIAlign [29] is used to align the input image and the feature maps properly. Second, the bounding box classification & regression, and mask segmentation are performed in parallel to predict the class, location, and profile of each object contained in bounding boxes. Third, we delete masks containing more than one component since cells are simply connected.

The DVS part generates probability distribution $p = (p_0, \ldots, p_K)$ over $K + 1$ classes ($p_0$ for background), bounding box regression offsets $t^k \in \mathbb{R}^4$ for $k = 1, \ldots, K$, and a binary mask $\hat{y} \in \mathbb{R}^{M \times N}$ of the ground truth class $k_{\text{gt}}$ for each RoI. Each RoI is labeled with a class $k_{\text{gt}}$, a bounding box offsets vector $v \in \mathbb{R}^4$, and a binary mask

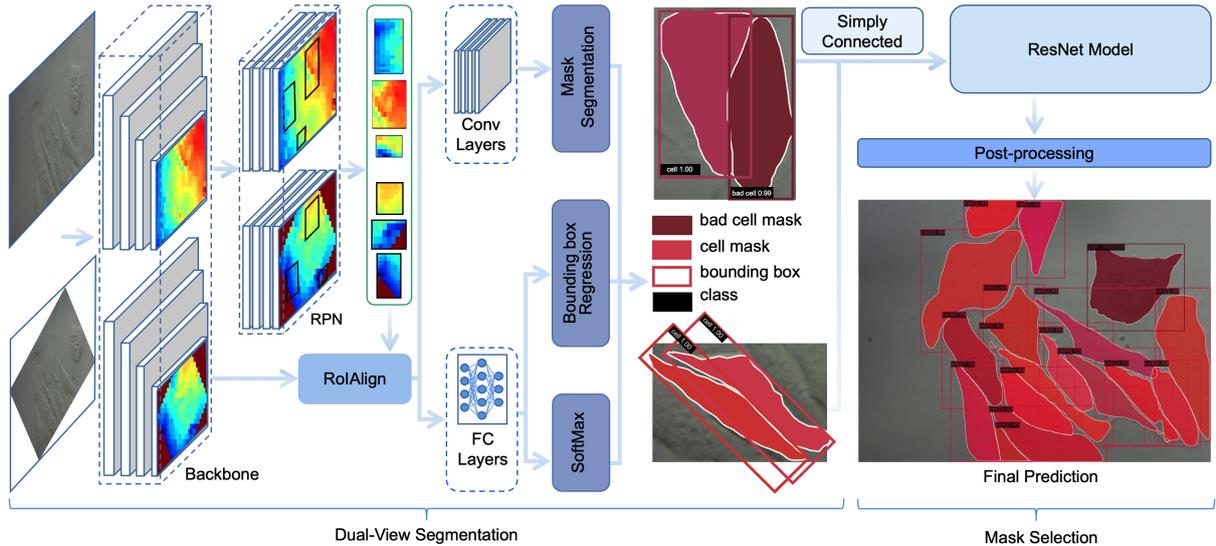

Fig. 3: Overview of our instance segmentation algorithm for unstained live adherent cells in DIC images, DVSISN. It consists of two parts, a DVS part (left) and an MS part (right). The DVS part takes a pair of original and rotated images as input and generates unfiltered bounding boxes and masks of identified cells as output. The middle visualizations show the bounding boxes and masks predicted by the DVS. Masks consisting of multiple pieces (that are not simply connected) are removed. After that, the MS part filters cell instances with high quality and generates the final prediction.

$y \in \mathbb{R}^{M \times N}$ in training using the multi-task loss as in the Mask R-CNN:

$$L = L_b + L_c + L_m, \quad (1)$$

where $L_b(k, t^k, v) = \mathbb{1}[k \geq 1] \sum_{i=1}^{4} d(t_i^k - v_i)$ accounts for the bounding box regression loss with

$$d(x) = \begin{cases} 0.5x^2, & \text{if } |x| < 1, \\ |x| - 0.5, & \text{otherwise}, \end{cases} \quad (2)$$

$L_c(p, k) = -\log p_k$ accounts for the classification loss of predictions of cell types, and $L_m$ is the average binary cross-entropy loss only defined for $k_{gt}$ of each RoI:

$$L_m(y, \hat{y}) = -\frac{\sum_{i=1}^{M} \sum_{j=1}^{N} y_{i,j} \log s(\hat{y}_{i,j}) + (1 - y_{i,j}) \log(1 - s(\hat{y}_{i,j}))}{MN}. \quad (3)$$

### 4.2 Mask Selection (MS)

The MS part consists of a ResNet classifier [48] and a post-processing module, as shown in Fig. 3 (right) and with details in Fig. 4. This part is responsible for removing unwanted bounding boxes generated by the DVS part, since feeding a pair of an original and rotated images into the DVS almost doubles the number of predicted bounding boxes, as each cell is often searched twice that leads to repeat detection of cells. Additionally, since the unsupervised non-maximum suppression (NMS) technique can efficiently remove duplicates only when cells are scattered, we relax its selection criteria by increasing the IoU threshold of NMS in DVS and add a supervised selection step, namely MS, to keep the predicated masks that are closest to the ground truth.

A cell mask $m$ produced by the DVS part is assigned a label $y_m = 1$ if it has the maximum IoU with the ground truth. Otherwise, the cell mask is assigned a label $y_m = 0$. Then these constructed cell masks and their binary labels are used to train a ResNet equipped with a cross-entropy loss to select appropriate masks. Finally, masks having the largest IoU (and also over 0.7) with other masks at "each spot" are preserved to prevent redundancies. As such, the MS is designed as a supervised selection step to keep the best mask predictions only.

## 5 Experimental Results

This section shows a comparison of our algorithm to the SOTA algorithms along with ablations on our dataset. Our dataset contains 520 images and is randomly partitioned into three parts: 312 images for training, 104 for validation, and 104 for testing. Six metrics [49] that calculate the average precision (AP) of bounding boxes and masks with different thresholds are used to report the performances of evaluated algorithms, as shown in Table 1. Most experiments were conducted on two NVIDIA 2080 Ti GPUs.

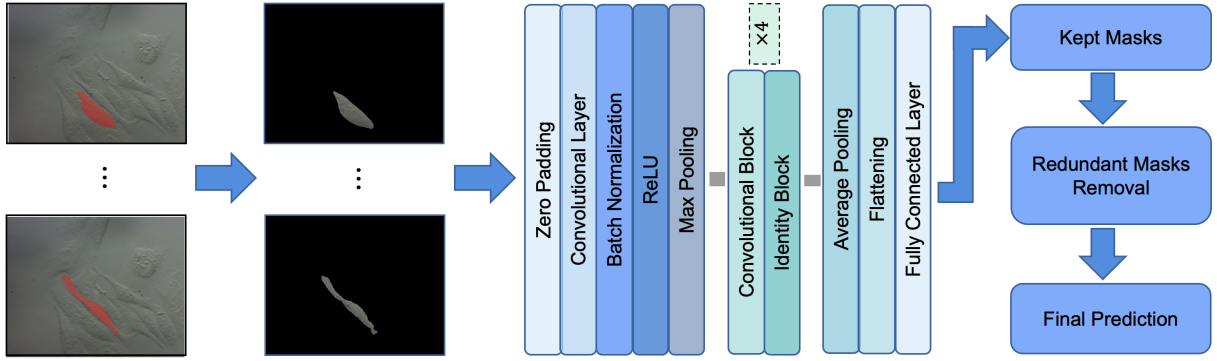

Fig. 4: Flow chart of the MS part. As shown on the left, masks generated by the DVS are used to crop the DIC images as the input of the MS. The middle part shows the structure of the ResNet-34 model that implements ResNet architecture in 3 parts. The first part uses $7 \times 7$ filters followed by a max-pooling layer to extract features. Then, 4 convolutional blocks and identity blocks are applied to use residual information, thus avoiding gradient vanishing. Finally, average pooling, flattening, and fully connected layers are used to decide if a mask will be kept. Then we delete redundant masks at "each spot" and produce the final instance segmentation of each DIC image.

Table 1: Average Precision Metrics for Object Detection and Instance Segmentation

| Metrics | Meaning |
| --- | --- |
| $AP^{bbox}$ | AP at IoU $= 0.50 : 0.05 : 0.95$ (primary challenge metric) for object detection, i.e., drawing bounding boxes of detected objects. |
| $AP^{bbox}_{0.50}$ | AP at IoU $= 0.50$ (PASCAL[a] VOC metric) for object detection. |
| $AP^{bbox}_{0.75}$ | AP at IoU $= 0.75$ (strict metric) for object detection. |
| $AP^{segm}$ | AP at IoU $= 0.50 : 0.05 : 0.95$ (primary challenge metric) for instance segmentation, i.e., generating individual masks of detected objects. |
| $AP^{segm}_{0.50}$ | AP at IoU $= 0.50$ (PASCAL VOC[b] metric) for instance segmentation. |
| $AP^{segm}_{0.75}$ | AP at IoU $= 0.75$ (strict metric) for instance segmentation. |

[a] PASCAL stands for pattern analysis, statistical modelling and computational learning [50].
[b] VOC stands for visual object classes [50].

## 5.1 Implementation Details

We implemented our algorithm based on the MMDetection toolbox [51] with the PyTorch framework [52]. DVSISN is trained in two stages. First, the backbone networks of the DVS are pre-trained on the COCO dataset [49] and tuned on our dataset. Second, The MS part is pre-trained on the ImageNet dataset [53], fine-tuned on the cell masks produced by the DVS, and constructed binary labels $y_m$ described in Section 4.2.

Data augmentation techniques, such as flipping, padding, and resizing, are used to increase training samples in DVS. We assign each GPU two input images and use RPN to generate RoIs. An RoI is regarded as positive if its IoU with a ground truth is over 0.7. Moreover, the RPN anchors are constructed by 5 aspect ratios 0.3, 0.5, 1, 2, 3, with a fixed scale 8, representing the length of an anchor's shortest side. ResNet-34 is used as the backbone of MS with a batch size of 32. We used stochastic gradient descent (SGD) with an initial learning rate of 0.05, a weight decay of 0.0001, a momentum of 0.9, and 500 iterations of warm-up.

## 5.2 Quantitative Results

The quantitative results of adherent cell instance segmentation are shown in Table 2. Mask R-CNN [29], Cascade R-CNN [41], Mask Scoring R-CNN [40], InstaBoost [30], and YOLACT [31] equipped with backbones ResNet-50 [48], ResNet-101 [48], and ResNeXt-101 [54] were used to compare against our algorithm.

Table 2: Quantitative Results of Adherent Cell Instance Segmentation

| Algorithm | Backbone | $AP^{bbox}$ | $AP^{bbox}_{0.50}$ | $AP^{bbox}_{0.75}$ | $AP^{segm}$ | $AP^{segm}_{0.50}$ | $AP^{segm}_{0.75}$ |
|---|---|---|---|---|---|---|---|
| Mask R-CNN [29] | ResNet-50 | 0.375 | 0.761 | 0.326 | 0.415 | 0.753 | 0.439 |
|  | ResNet-101 | 0.410 | 0.771 | 0.389 | 0.414 | 0.772 | 0.486 |
|  | ResNeXt-101 | 0.447 | 0.768 | 0.468 | 0.431 | 0.776 | 0.454 |
| Cascade R-CNN [41] | ResNet-50 | 0.455 | *0.776* | 0.475 | 0.437 | 0.778 | 0.483 |
|  | ResNeXt-101 | *0.459* | 0.764 | *0.497* | 0.443 | 0.778 | 0.492 |
| Mask Scoring R-CNN [40] | ResNet-50 | 0.437 | 0.774 | 0.438 | *0.449* | *0.793* | *0.496* |
|  | ResNeXt-101 | 0.438 | 0.772 | 0.467 | 0.440 | 0.770 | 0.485 |
| InstaBoost [30] | ResNet-50 | 0.429 | 0.749 | 0.436 | 0.419 | 0.756 | 0.443 |
|  | ResNeXt-101 | 0.434 | 0.739 | 0.462 | 0.429 | 0.768 | 0.467 |
| YOLACT [31] | ResNet-50 | 0.343 | 0.688 | 0.291 | 0.329 | 0.651 | 0.293 |
|  | ResNet-101 | 0.351 | 0.709 | 0.300 | 0.335 | 0.674 | 0.316 |
| DVSISN | ResNet-50 | <u>0.609</u> | <u>0.955</u> | <u>0.648</u> | <u>0.549</u> | <u>0.889</u> | <u>0.632</u> |
|  | ResNet-101 | **0.634** | **0.968** | **0.686** | **0.555** | **0.892** | **0.647** |

Bold and underlined values indicate the best and the second-best performances. Italic values are the best performances reported by competitive algorithms.

It can be observed clearly that the DVSISN outperforms all counterparts in terms of all six metrics. Its $AP^{bbox}$ is over 15% better than its closest counterpart (Cascade R-CNN [41]) on all tested backbones. Additionally, its $AP^{bbox}_{0.50}$ is 0.968, implying that almost all cells are successfully detected, leading to a substantial improvement on the more strict metric $AP^{bbox}_{0.75}$. Thanks to the nearly perfect cell detection, DVSISN's instance segmentation performs nicely; DVSISN wins 13% more than the second-best algorithm (Mask Scoring R-CNN [40]) even on the most critical metric $AP^{segm}_{0.75}$.

In contrast, the $AP^{bbox}$ and $AP^{segm}$ of all the other algorithms are lower than 0.5, regardless of ResNet backbone choices, failing our expectations at the beginning of this study. Instead, DVSISN outperforms all other algorithms in all six metrics by over 10%. We can conclude that adopting the DVS and MS improves the performance of DVSISN remarkably.

## 5.3 Qualitative Results

The qualitative results of adherent cell instance segmentation are displayed in Fig. 5. At first glance, it seems that these algorithms (Mask R-CNN [29], Cascade R-CNN [41], Mask Scoring R-CNN [40], InstaBoost [30], YOLACT [31]) can identify individual cells relatively well, but in fact, their inference details are not satisfactory. For example, as shown in Figs. 5(c-1) to 5(g-1), a few titled cells were always neglected, especially when cells were densely distributed. However, as shown in Figs. 5(c-2) to 5(g-2), there existed fragmented mask predictions, implying that a cell's mask was mispredicted even if the cell was detected correctly. Furthermore, as shown in Figs. 5(c-3) to 5(g-4), overlapping mask predictions can be observed, which means that the NMS technique did not successfully filter a few unwanted masks. Last but not least, as shown in Figs. 5(c-5) to 5(g-5), an obvious but tilted cell in the upper left corner was neglected even though cells in the input image Fig. 5(a-5) are not densely distributed, meaning that the detection accuracy of these existing algorithms still has much room for improvement. Compared to these SOTA algorithms, our DVSISN demonstrates remarkable improvement. It can accurately detect cells in both sparsely and densely distributed situations. Based on accurate detection, mask predictions can be achieved.

## 5.4 Ablation Study

The ablation study for different backbones with and without DVS or MS is listed in Table 3. In Table 3, DVSISN[†] is equipped with DVS only, while DVSISN[‡] is equipped with MS only.

The performance of DVSISN[†] indicates that the DVS improves $AP^{bbox}$ and $AP^{segm}$ from 3% to 5% compared to the Mask R-CNN. Although the $AP^{bbox}_{0.50}$ and $AP^{segm}_{0.50}$ of the DVSISN[†] approximate those of the Mask R-CNN, $AP^{bbox}_{0.75}$ and $AP^{segm}_{0.75}$ of the DVSISN[†] wins by a large margin, especially with a ResNet-50 as the backbone (around 10%), implying that DVSISN[†] makes better high-quality predictions than the Mask R-CNN.

The performance of DVSISN[‡] shows that the MS can efficiently select high-quality masks in a supervised way, thus indirectly improving the quality of their corresponding bounding boxes. DVSISN[‡] equipped with a ResNet-50

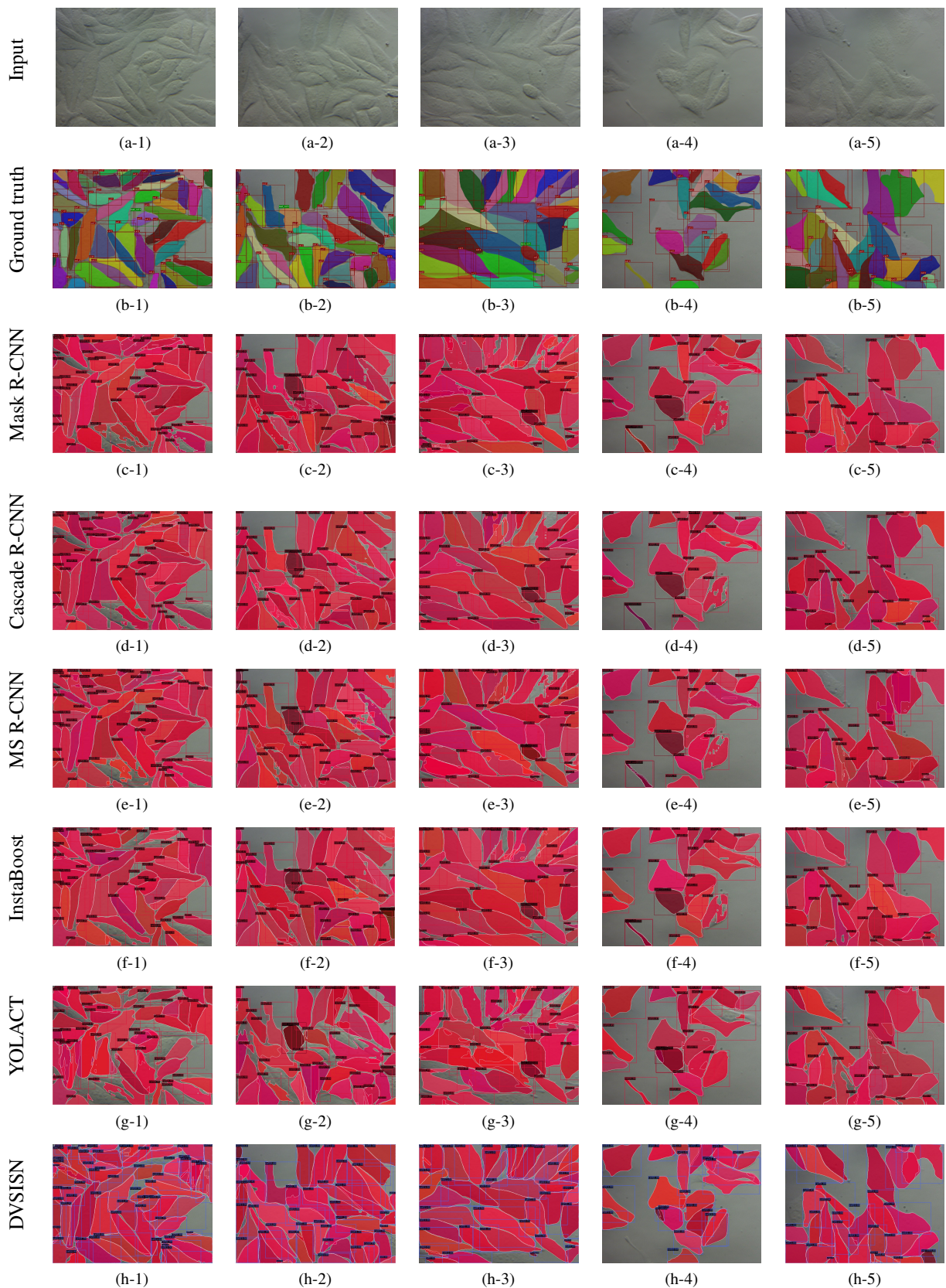

Fig. 5: Qualitative inference results of adherent cell images. Our algorithm outperforms the other counterparts.

backbone outperforms Mask R-CNN in both $AP^{bbox}$ and $AP^{segm}$, but the one with ResNet-101 only surpasses Mask R-CNN in $AP^{bbox}$ slightly while failing in $AP^{segm}$.
9

Table 3: Quantitative Results of the Ablation Study

| Algorithm | Backbone | DVS | MS | $AP^{bbox}$ | $AP^{bbox}_{0.50}$ | $AP^{bbox}_{0.75}$ | $AP^{segm}$ | $AP^{segm}_{0.50}$ | $AP^{segm}_{0.75}$ |
|---|---|---|---|---|---|---|---|---|---|
| Mask R-CNN [29] | ResNet-50 | ✗ | ✗ | 0.375 | 0.761 | 0.326 | 0.415 | 0.753 | 0.439 |
|  | ResNet-101 | ✗ | ✗ | 0.410 | 0.771 | 0.389 | 0.414 | 0.772 | 0.486 |
| DVSISN† | ResNet-50 | ✓ | ✗ | 0.426 | 0.750 | 0.448 | 0.449 | 0.732 | 0.533 |
|  | ResNet-101 | ✓ | ✗ | 0.450 | 0.799 | 0.469 | 0.463 | 0.794 | 0.520 |
| DVSISN‡ | ResNet-50 | ✗ | ✓ | 0.499 | 0.761 | 0.529 | 0.455 | 0.739 | 0.513 |
|  | ResNet-101 | ✗ | ✓ | 0.450 | 0.830 | 0.424 | 0.413 | 0.788 | 0.466 |
| DVSISN | ResNet-50 | ✓ | ✓ | 0.609 | 0.955 | 0.648 | 0.549 | 0.889 | 0.632 |
|  | ResNet-101 | ✓ | ✓ | 0.634 | 0.968 | 0.686 | 0.555 | 0.892 | 0.647 |

ResNet-50 and ResNet-101 are used as backbones in the experiments. Mask R-CNN is used as a benchmark. DVSISN† only adopts the DVS. DVSISN‡ only adopts the MS. DVSISN adopts both DVS and MS.

A full DVSISN equipped with both DVS and MS performs better than DVSISN† and DVSISN‡. It is because the DVS generates sufficient cell-aligning bounding boxes, and then the MS keeps only bounding boxes associated with well-predicted masks. In a nutshell, using DVS or MS alone brings a slight improvement, but using them together brings remarkable advancement.

# 6 Conclusion

In this study, we developed a new algorithm called DVSISN for segmenting unstained live adherent cells in DIC images. Experimental results demonstrate that the DVSISN outperforms major SOTA algorithms by a large margin, approximately 10% to 20%, in terms of $AP^{bbox}$ and $AP^{segm}$. Such an advantage can be attributed to two novel methods—DVS and MS—that take combinations of original and rotated views as input to capture cell instances as much as possible and select the finest instances in a supervised way. Ablation studies further confirmed that the DVS could squeeze bounding boxes to better align with cell instances of various orientations, and the MS can keep high-quality masks that improve the AP of masks, thus indirectly improving the quality of bounding boxes. In short, our DVSISN is an accurate and robust algorithm for adherent cell segmentation.

We plan to integrate the DVSISN into our cell micromanipulation system [7] to conduct intracellular deliveries to investigate biological and biophysical reactions [55, 56]. Meanwhile, we plan to implicitly merge the DVS part into the training of RPN to reduce the computational cost in training [57]. We also plan to test quadrilateral bounding boxes to check final performance [58].

# Acknowledgment

This work was partially conducted by using the computational facilities, CityU Burgundy, managed and provided by the Computing Services Center at the City University of Hong Kong.